\documentclass[conference]{IEEEtran}
\IEEEoverridecommandlockouts

\usepackage{graphicx}
\usepackage{amsmath,amssymb,amsfonts}
\usepackage{algorithmic}
\usepackage{textcomp}
\usepackage{times,epsfig,graphicx}
\usepackage{algorithm,algorithmic}
\usepackage[usenames]{color}

\usepackage{amssymb,amsmath}
\usepackage[latin1]{inputenc}
\usepackage{fancyhdr}
\usepackage{verbatim}
\usepackage{tabularx}
\usepackage{hyperref}
\usepackage{xspace}
\usepackage{bm}

\newlength{\smallimage}
        \definecolor{rel}{rgb}{.1,.6,.2}
        \definecolor{nrl}{rgb}{1,1,1}
        \definecolor{qim}{rgb}{1,1,1}

\makeatletter

\DeclareRobustCommand\onedot{\futurelet\@let@token\@onedot}
\def\@onedot{\ifx\@let@token.\else.\null\fi\xspace}

\def\be{\begin{equation}}
\def\ee{\end{equation}}
\def\bea{\begin{eqnarray}}
\def\eea{\end{eqnarray}}
\def\ben{\begin{eqnarray*}}
\def\een{\end{eqnarray*}}

\def\bi{\begin{itemize}}
\def\ei{\end{itemize}}

\newcommand{\bt}[1]{\begin{tabular}{#1}}
\newcommand{\et}{\end{tabular}}
\newcommand{\ba}[1]{\begin{array}{#1}}
\newcommand{\ea}{\end{array}}
\newcommand{\ul}[1]{\underline{#1}}

\def\<{\langle}
\def\>{\rangle}

\newcommand{\bfx}{{\bf x}}
\newcommand{\bfy}{{\bf y}}
\newcommand{\bfw}{{\bf w}}

\newcommand{\bfv}{{\bf v}}

\newcommand{\bfz}{{\bf z}}

\newcommand{\bfV}{{\bf V}}

\newcommand{\bfW}{{\bf W}}

\newcommand{\bfb}{{\bf b}}
\newcommand{\bfd}{{\bf d}}

\newcommand{\hide}[1]{}

\def\BibTeX{{\rm B\kern-.05em{\sc i\kern-.025em b}\kern-.08em
    T\kern-.1667em\lower.7ex\hbox{E}\kern-.125emX}}
    
\begin{document}

\title{\LARGE \bf Learning Common Representation from RGB and Depth Images}

\author{Giorgio Giannone$^{1}$ and Boris Chidlovskii$^{2}$  
\thanks{$^{1}$Giorgio Giannone is with Sapienza University of Rome, Italy, {\tt\small giorgio.c.giannone@gmail.com}}%
\thanks{$^{2}$Boris Chidlovskii is with Naver Labs Europe,  France, {\tt\small boris.chidlovskii@naverlabs.com}}%
}

%\author{Boris Chidlovskii and Giorgio Giannone}
%\institute{Naver Labs Europe, Sapienza University Rome} %Paper ID \ECCV18SubNumber}
%\author[1]{Giorgio Giannone}
%\author[2]{Boris Chidlovskii}
%\affil[1]{Sapienza University of Rome, Naver Labs Europe}
%\affil[2]{Naver Labs Europe}
%\institute{}

\maketitle
%%%%%%%%%%%%%%%%%%%%%%%%%%%%%%%%%%%%%%%%%%%%%%%%%%%%%%

\begin{abstract}
We propose a new deep learning architecture for the tasks of semantic segmentation and depth prediction from RGB-D images. We revise the state of art based on the RGB and depth feature fusion, where both modalities are assumed to be available at train and test time. We propose a new architecture where the feature fusion is replaced with a common deep representation. Combined with an encoder-decoder type of the network, %the state of art network for semantic segmentation, 
the architecture can jointly learn models for semantic segmentation and depth estimation based on their common representation. 
This representation, inspired by multi-view learning, offers several important advantages, such as using one modality available at test time  to reconstruct the missing modality. In the RGB-D case, this enables the cross-modality scenarios, such as using depth data for semantically segmentation and the RGB images for depth estimation. We demonstrate the effectiveness of the proposed network on two publicly available RGB-D datasets. The experimental results show that the proposed method works well in both semantic segmentation and depth estimation tasks.
%Evaluation on NYU2 dataset (standard RGB-D dataset) show an important mIoU increase with respect to the baseline. %Robustness to the failure of some sensors.
\end{abstract}

% TO DO list: fix modality vs view
% fix class vs categories
% fix common representation vs CRL, learning vs embedding
% define x^r d^r in Figure 1 OK
% fix dilation vs upsampling OK
%=================================================
\section{INTRODUCTION}
\label{sec:intro}

Visual scene understanding is a critical capability enabling robots to act in their working environment. Modern robots and autonomous vehicles are equipped with many, often complementary sensing technologies. Multiple sensors aim to satisfy the need for the redundancy and robustness  
critical for achieving the human level of the driving safety.
%/fault tolerance

%Among the many technologies which make autonomous vehicles possible is a combination of sensors and actuators, sophisticated algorithms, and powerful processors to execute software. The sensors and actuators in an autonomous vehicle fall into three broad categories: 1) navigation and guidance (where you are, where you want to be, how to get there); 2) driving and safety (directing the vehicle, making sure it vehicle acts properly under all circumstances, and follows the rules of the road); and 3) performance (managing the car's basic internal systems).

The most frequent case is RGB-D cameras collecting color and depth information for different computer vision tasks~\cite{cityscape16,janai17,song15}.
%, such as Microsoft Kinect. Depth data can help conventional RGB image recognition by providing additional information to better understand the spatial layout of the objects and regions in the scene. 
%The depth channel can be easily captured with low cost RGB-D sensors.
%FuseNet~\cite{hazirbas16} addressed the problem of semantic labeling of indoor scenes on RGB-D data. 
%With the availability of RGB-D cameras, it is expected that additional depth measurement will improve the accuracy. Here we investigate a solution how to incorporate complementary depth information into a semantic segmentation framework by making use of convolutional neural networks (CNNs).
%we address the problem of indoor scene understanding assuming that both RGB and depth information simultaneously available. This problem is rather crucial in many perceptual applications including robotics. 
As information collected by the depth camera is complementary to RGB images,
%Utilizing depth additional to the appearance information (i.e. RGB) 
the depth can potentially help decode structural information of the scene and improve the performance on such tasks as object detection and semantic segmentation~\cite{song15}. 
% In general object classes can be recognized based on their color and texture attributes. 
%However, the auxiliary depth may reduce the uncertainty of the segmentation of objects having similar appearance information. 
%Couprie et al .[9] observed that the segmentation of classes having similar depth, appearance and location is improved by making use of the depth information too, but it is better to use only RGB information to recognize object classes containing high variability of their depth values. 
%Therefore, the optimal way to fuse RGB and depth information has been left an open question.

% ==== BEFORE DEEP ====
%RGB-D processing before deep learning and CNN

%RGB-D scene recognition has attracted increasingly attention due to the rapid development of depth sensors and their wide application scenarios. While many research has been conducted, most work used hand-crafted features which are difficult to capture high-level semantic structures.~\cite{cityscape16}.

% ==== RECENT SUCCESS of DL ===
%With the explosive growth of image datasets and the 

%With the successful development of convolutional neural networks (CNNs) the performance of RGB image recognition has been dramatically improved.

%\paragraph{FuseNet and GRB-D} \cite{hazirbas16}
Development of convolutional neural networks (CNNs) boosted the performance of the image classification. Recently, their success has been replicated on  object detection and semantic segmentation tasks. 
%CNNs have been shown to be powerful visual models that yields hierarchies of features. 
The key contribution of CNN models lies in their ability to model complex visual scenes. Current CNN-based approaches provide the state-of-the-art performance in semantic segmentation benchmarks~\cite{deeplab2,garcia17}.
%In contrast to CNN models, by applying hand-crafted features one can generally achieve rather limited accuracy.

%===== DEEP RGB-D %\paragraph{Fusenet, continued}
When RGB images are completed with depth information, the straightforward idea is to incorporate depth information into a semantic segmentation framework. Different methods have been developed including deep features pooling, dense feature, multi-scale fusion, etc.~\cite{eigen14,eitel15,gupta14,ladicky14,silberman12}. Most recent methods, like FuseNet~\cite{hazirbas16,jiang17}, use an encoder-decoder architecture, where the encoder part is composed of two branches that simultaneously extract features from RGB and depth images and {\it fuse} depth features into the RGB feature maps.
% as the network goes deeper. In FuseNet, two different ways for fusion of the RGB and depth channels were tested. 
%We also analyze the proposed network architectures, referred to as dense and sparse fusion, in terms of the level of fusion.
%\item We experimentally show that our proposed method is successfully able to fuse RGB and depth information for semantic segmentation also on cluttered indoor scenes. Moreover, our method achieves competitive results with state-of-the-art methods in terms of segmentation accuracy evaluated on the challenging SUN RGB-D dataset [10].
%
%===== Multi-view approach to RGB-D
Moreover, training individual RGB and depth views has been replaced with the joint learning.
%~\cite{ma17} proposed a novel approach to object-class segmentation from multiple RGB-D views using deep learning. 
%We train a deep neural network to predict object-class semantics that is consistent from several view points in a semi-supervised way. 
%At test time, 
It was shown that the semantics predictions of jointly learned network can be fused more consistently than predictions of a network trained on individual views~\cite{ma17}. 
%%In a jointly learned network, 
%base our network architecture on a recent single-view deep learning approach to RGB and depth fusion for semantic object-class segmentation 
%%the objective function to minimize enhances the individual view losses on multiple scales.
%and enhance it with multi-scale loss minimization. 
%We obtain the camera trajectory using RGB-D SLAM and warp the predictions of RGB-D images into ground-truth annotated frames in order to enforce multi-view consistency during training. At test time, predictions from multiple views are fused into keyframes. 

%Other methods exploit the complementarity between the RGB and depth data, they explicitly analyze and enforce consistency between the two channels during training and testing~\cite{ma17}.
% They enforced such a multi-view consistency during the training and fuse features over multiple views.
% outperforms single-view baselines. % on the NYUDv2 benchmark for semantic segmentation. 
%Our end-to-end trained network achieves state-of-the-art performance on the NYUDv2 dataset in single-view segmentation as well as multi-view semantic fusion.
%--------------------------
\begin{figure}[ht]
\centering{\includegraphics[width=0.5\textwidth]{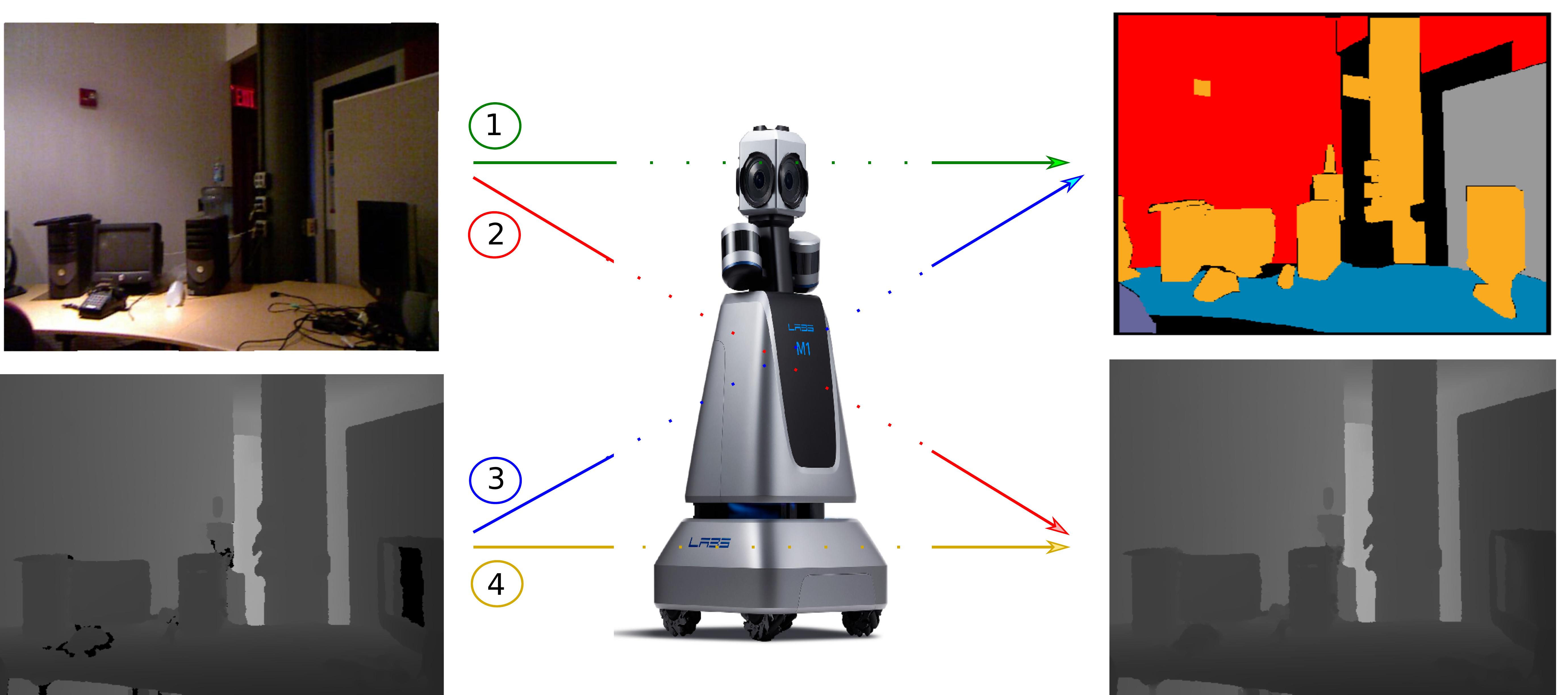}}
\caption{Scenarios for RGB-D data include semantic segmentation from RGB (1), depth (3) or both (1+3), depth prediction from RGB (2) and depth completion from depth (4).}
\label{fig:scenarios}
\end{figure}
%-------------------------

%===============================================
%\paragraph{Our proposal} %\label{ssec:novelty}
In this paper, we propose a new deep learning architecture for the tasks of semantic segmentation and depth estimation from RGB-D images (see Figure~\ref{fig:scenarios} for different scenarios).
%To our best knowledge, our approach differs from the state of art methods in several important issues, as follows:
%\begin{enumerate}
%\item %We propose a new deep architecture %for a joint learning of semantic segmentation and depth estimation; it is 
%able to work with both RGB and RGB-D images as input. 
Usually, these tasks are addressed separately, with a special design for semantic segmentation~\cite{hazirbas16,ma17} or depth prediction~\cite{eigen14}. We develop a unifying framework capable to cope with either task. 
%Moreover it allows to process more scenarios than in the state of art approaches. The network can work with RGB-D data as input, trains its network for semantic segmentation, depth estimation or both. 
% and cope with different scenario, when all or some views are available at the test time.

We adopt the multi-view approach to RGB-D data, where RGB and depth are complementary sources of information about a visual scene. All existing methods, whether they train the view models independently or jointly, %assume the dominance of the RGB view, eventually complemented with the depth. And they all 
adopt the fusion-based representation. The feature fusion takes  benefit from the view complementarity to reduce the uncertainty of segmentation and labelling. The fusion-based approaches however require both views to be available at test time. 

We revise the fusion-based approach and replace it with the common representation~\cite{ngiam11}. Adopting the principle of the common representation gives a number of benefits well studied in the multi-view learning~\cite{baltrusaitis17}. First, it allows to obtain the common representation from one view and then reconstruct all other views. It can accomplish the task when one view is unavailable due to technical or other reasons, thus increasing the robustness and fault-tolerance of the system. Working with one-view data at test time enables {\it cross-view scenarios} rarely addressed in the state of art. In semantic segmentation, when the RGB view is unavailable, the depth view can be used to obtain the common representation and accomplish the semantic segmentation task. And vice-versa, in the case of a depth estimation, the common representation allows to use the RGB view to reconstruct the depth of a scene or an object.

Second, the common representation is the central component allowing to deploy the same architecture for both RGB-D tasks. 
%In case of semantic segmentation, 
Representation common to RGB and depth 
% and the depth estimation using the GRB images. 
allows to enforce the consistency between the views and improve the segmentation quality and depth estimation accuracy.  
%Use the depth information to validate/make consistency and complementary of the from RGB channel. 
%When the semantic segmentation is coupled with the depth estimation, it enforces the mutual consistency between the views for the multi-tasking. 
%The later is a strong alternative to the recent trend, with a specially designed networks for each task.

Third, the proposed architecture 
%allows exploring additional single view data~\cite{ngiam11} at train time. Finally, this 
encourages a higher modularity of the deep network.  Our proposal combines the state of art components, the encoder-decoder networks for semantic segmentation and a multi-view autoencoder for the common representation. The system can then benefit from any progress in individual components. The modularity allows to upgrade a component without changing the entire system, training and optimization routines. 

The remainder of the paper is organized as follows. In Section~\ref{sec:soa}, we review the state of art of semantic segmentation and depth estimation for RGB-D images. In Section~\ref{sec:arch}, we introduce the multi-view deep architecture and describe in details each component, the two-stage training and optimization. Section~\ref{sec:evaluation} reports results of evaluating the network on two public RGB-D datasets; it also discusses some open questions. Section~\ref{sec:conclusion} concludes the paper.
%We demonstrate the effectiveness of the proposed network on two publicly available RGB-D datasets. the experimental results show that the proposed method work well in both semantic segmentation and depth estimation tasks.

%--------------------------
\begin{figure*}[ht]
\centering{\includegraphics[width=0.75\textwidth]{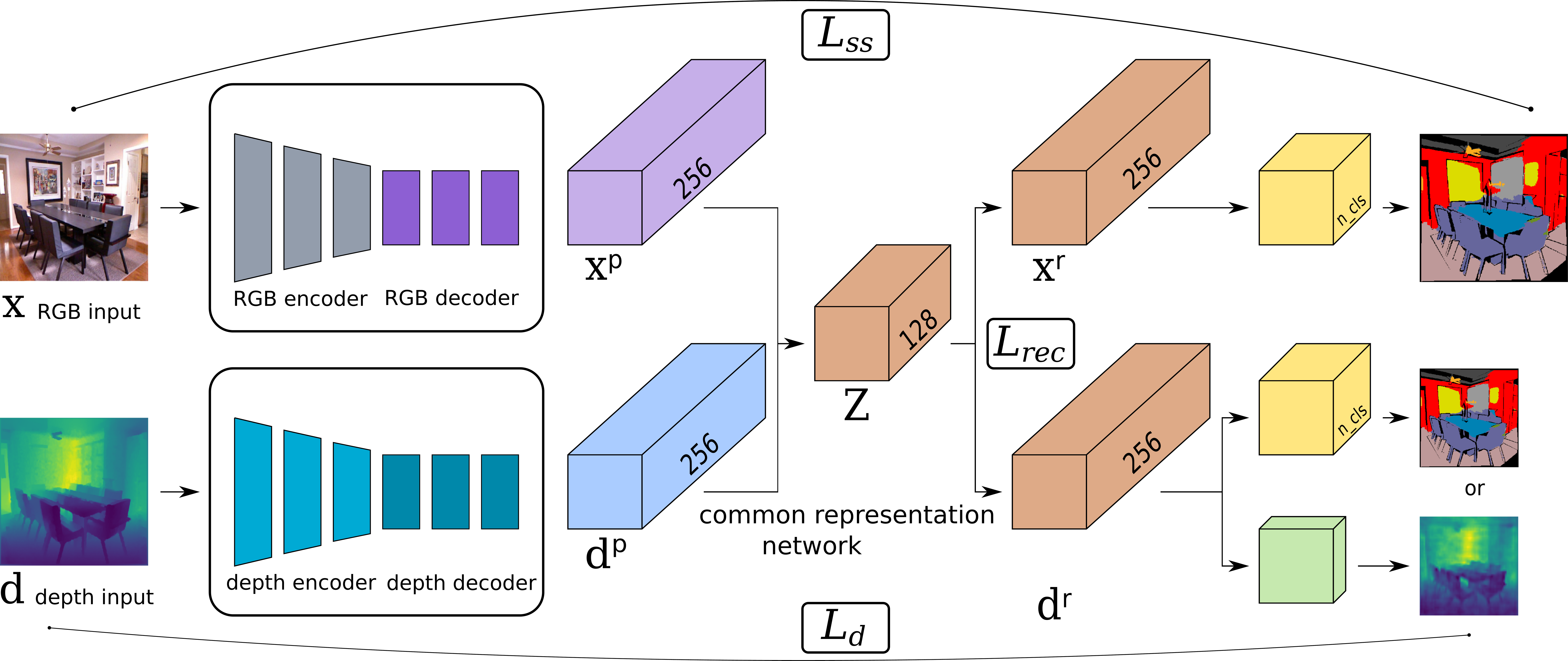}}
\caption{The architecture is composed of two encoder-decoder networks for RGB and depth images and the common representation network. Depending on the setting, the depth network is trained with the segmentation labels or depth ground true values.}
\label{fig:arch}
\end{figure*}
%-------------------------

%=============================================
%------------ STATE OF ART ----------------
%=================================================
\section{STATE OF ART}
\label{sec:soa}

{\it Depth representation.}
Depth information is rarely used in any segmentation network as raw data, most methods use so called {\it HHA representation} of the depth~\cite{gupta14}. This representation consists of three channels: disparity, height of the pixels and the angle between normals and the gravity vector based on the estimated ground floor. 
%By making use of the HHA representation, a superficial improvement was achieved in terms of segmentation accuracy. On the other hand, the information retrieved only from the RGB channels still dominates the HHA representation. As we shall see in Section 4, the HHA representation does not hold more information than the depth itself. %Furthermore, computing HHA representation requires high computational cost. 
The color code provided by HHA helps visualize depth information; it can reveal some patterns that resemble RGB patterns. 
%In [Wang et al., 2016] RGB and depth CNNs are obtained by fine tuning CNN models pretrained on large scale RGB datasets, then the resulting

%--------------------------
{\it Semantic segmentation and depth estimation.}
These two fundamental tasks for RGB-D images 
%have been previously addressed separately. %However they both used similar  convolution NN architectures.
%As the semantic segmentation and depth estimation 
are strongly correlated and mutually beneficial, and
most efforts were on putting both views in one architecture. In particular, with the success of CNN architectures, many methods aimed to inject the depth information into the semantic segmentation network~\cite{eitel15,hazirbas16,ladicky14,jiang17,ma17,valada17}.

%-------------------------------------------
%\paragraph{Feature fusion and joint learning.} %Depth estimation and completion}
%-------------------------------------------
%\paragraph{Joint learning.}
Ladicky at al.~\cite{ladicky14} were first to replace single-view depth estimation and semantic segmentation by a joint training model. %for semantic segmentation and path estimation. 
They considered both semantic label loss and depth label loss when learning a classifier.
%The limitations of current state-of-the-art methods for  are closely tied to the property of perspective geometry, that the perceived size of the objects scales inversely with the distance. 
Using properties of perspective geometry, they reduced the learning of a pixel-wise depth classifier to a simpler classifier predicting %only the likelihood of a pixel being at 
one of fixed canonical depth values~\cite{ladicky14}.
%The likelihoods for any other depths can be obtained by applying the same classifier after appropriate image manipulations. 
%Such transformation of the problem to the canonical depth removes the training data bias towards certain depths and the effect of perspective. 
%The approach can be straight-forwardly generalized to multiple semantic classes, improving both depth estimation and semantic segmentation performance by directly targeting the weaknesses of independent approaches. 
%Conditioning the semantic label on the depth provides a way to align the data to their physical scale, allowing to learn a more discriminative classifier. 
%By conditioning depth on the semantic class, the classifier was able to distinguish between ambiguous predictions. % of the otherwise ill-posed problem. 
%We tested our algorithm on the KITTI road scene dataset and NYU2 indoor dataset and obtained obtained results that significantly outperform current state-of-the-art in both single-view depth and semantic segmentation domain.
%They also used local regions with hand-crafted features for predictions, able to generate very coarse depth and semantic maps.

%------------------------------
Two separate CNN processing streams, one for each modality, were proposed by Eitel at al.~\cite{eitel15}; %{\it Eitel15 %"Multi-modal deep learning for robust RGB-D object recognition"}.
%Robust object recognition is a crucial ingredient of many, if not all, real-world robotics applications. This paper leverages recent progress on Convolutional Neural Networks (CNNs) and proposes a novel RGB-D architecture for object recognition. 
%was first to 
they are consecutively combined in a late fusion network. 
%We focus on learning with imperfect sensor data, a typical problem in real-world robotics tasks. 
%For accurate learning, we 
The method also introduced a multi-stage training methodology %and two crucial ingredients 
for handling depth data with CNNs. %The first, an effective encoding of depth information for CNNs that enables learning without the need for large depth datasets. 
%The second, a data augmentation scheme for robust learning with depth images by corrupting them with realistic noise patterns. We present state-of-the-art results on the RGB-D object dataset and show recognition in challenging RGB-D real-world noisy settings. 
It used the HHA representation of depth and the data augmentation scheme for robust learning with depth image.%, by corrupting them with realistic noise patterns.

%\cite{wang15} %{\it (wang15) }
%Using the correlation between the depth estimation and semantic segmentation are two fundamental problems in image understanding. 
%As the semantic segmentation and depth estimation are strongly correlated and mutually beneficial, 
%they are usually solved separately or sequentially. 
%Motivated by the complementary properties of the two tasks, 
A unified framework for joint depth and semantic prediction was proposed by Wang at al.~\cite{wang15}. Given an image, they first use a trained CNN to jointly predict a global layout composed of pixel-wise depth values and semantic labels. %By allowing interactions between the depth and semantic information, 
The joint network showed to provide more accurate depth prediction than a state-of-the-art CNN trained solely for depth prediction. To further obtain fine-level details, the image is decomposed into local segments for region-level depth and semantic prediction. 
By considering RGB and depth channels as multi-view data,~\cite{ma17} 
%We train a deep neural network to 
%predict object-class predictions by these two views are expected to be consistent
enforced the multi-view consistency during training and testing. 
%from several view points in a semi-supervised way.
At test time, the semantic predictions of the network are fused more consistently than predictions of a network trained on individual views. The network architecture uses a recent single-view deep learning approach to RGB and depth fusion and enhances it with multi-scale loss minimization. 
%We obtain the camera trajectory using RGB-D SLAM and warp the predictions of RGB-D images into ground-truth annotated frames in order to enforce multi-view consistency during training. 
%At test time, predictions from multiple views are fused into keyframes. 

%We evaluate the benefit of multi-view consistency training and demonstrate that pooling of deep features and fusion over multiple views outperforms single-view baselines on the NYUDv2 benchmark for semantic segmentation. Our end-to-end trained network achieves state-of-the-art performance on the NYUDv2 dataset in single-view segmentation as well as multi-view semantic fusion.

FuseNet~\cite{hazirbas16} %proposed a different way to %addressed the problem of semantic labeling of indoor scenes on RGB-D data. With the availability of RGB-D cameras, it is expected that additional depth measurement will improve the accuracy. Here we investigate a solution how to 
%incorporate complementary depth information into a semantic segmentation framework.
% by making use of convolutional neural networks (CNNs. 
%Recently encoder-decoder type fully convolutional CNN architectures have achieved a great success in the field of semantic segmentation. %Motivated by this observation 
%It 
developed an encoder-decoder type network, where the encoder part is composed of two branches of networks that simultaneously extract features from RGB and depth images and fuse depth features into the RGB feature maps as the network goes deeper. %Comprehensive experimental evaluations demonstrate that the proposed fusion-based architecture achieves competitive results with the state-of-the-art methods on the challenging SUN RGB-D benchmark.
%multi-view consistency and complementarity.

%\cite{jiang17} %"Incorporating Depth into both CNN and CRF for Indoor Semantic Segmentation". To improve segmentation performance, a novel neural network architecture (termed DFCN-DCRF) is proposed, which combines an RGB-D fully convolutional neural network (DFCN) with a depth-sensitive fully-connected conditional random field (DCRF). 
Although most of the above methods apply the late fusion, it is also possible to fuse depth information into the early layers of fully convolutional neural network~\cite{jiang17}. Coupled with the dilated convolution for later contextual reasoning, it combines a depth-sensitive fully-connected CRF with the previous convolution layers to refine the preliminary result. 
%Comparative experiments show that the proposed DFCN-DCRF achieves competitive performance compared with state-of-the-art methods.

%------------------------------
{\it Depth Completion.}
The problem of completing the depth channel of an RGB-D image has been addressed in~\cite{zhang18}. Indeed, it often the case that commodity-grade depth cameras fail to sense depth for bright, transparent, and distant surfaces thus leaving entire holes in the depth images. They train a deep network that takes an RGB image as input and predicts dense surface normals and occlusion boundaries. Those predictions are then combined with raw depth observations provided by the RGB-D camera to solve for depths for all pixels, including those missing in the original observation. 
\subsection{Multi-view learning}
\label{ssec:multi-view}

In the previous section we reviewed different ways to fuse RGB and depth features.
Meanwhile, there exist alternative representations for multi-view data~\cite{baltrusaitis17}. One such alternative, the {\it common representation learning} (CRL), tries to embed different views of the data in a common subspace~\cite{baltrusaitis17}. It allows to obtain a common representation from one view and use it to reconstruct other views.

Two complementary approaches to CRL are based on canonical correlation analysis (CCA) and multi-modal autoencoders (MAE). CCA based approaches learn a joint representation by maximizing correlation of the views when projected into the common subspace. 
%Canonical Correlation Analysis (CCA) is a commonly used tool for learning common representations for two-view data~\cite{dhillon11}). 
%By definition, CCA aims to produce correlated common representations %, but it suffers from some drawbacks. 
%First, 
%Kernelized versions of CCA replaced the linear projection by a non-linear one, 
%however KCCA remains hardly scalable to large datasets. %it is not easily scalable to very large datasets, in 
%Of course, there are some approaches which try to make CCA scalable (for example,~\cite{lu14}), but such scalability comes at the cost of performance. Further, since CCA does not explicitly focus on reconstruction, reconstructing one view from the other might result in low quality reconstruction. 
%Finally, CCA cannot benefit from additional non-parallel, single-view data. 
%This puts it at a severe disadvantage in several real world situations, where in addition to some parallel two-view data, abundant single view data is available for one or both views. 
%and a deep version of CCA has been proposed~\cite{andrew13} to replace a explicit kernel with a deep non-linear projection. 
%; such a embedding is learing by appliyng SGD on small data batches~\cite{wang15}.
%
%----- AE -----------------
%Autoencoder based methods learn a common representation by minimizing the error of reconstructing the two views. 
% For example, while CCA based approaches outperform AE based approaches for the task of transfer learning, they are not as scalable as the latter.
Second approach to embed two views is based on multi-modal autoencoders (MAEs) \cite{ngiam11}. 
%The multimodal autoencoders  have been proposed to learn a deep common representation for two views. 
The idea is to train an autoencoder able to perform two kinds of reconstruction. Given one view, the model learns both self-reconstruction and cross-reconstruction (reconstruction of the other view). %This makes the representation learned to be predictive of each other. %and reconstruct using the common representation.

%However, it should be noticed that the MAE does not get any explicit learning signal encouraging it to share the capacity of its common hidden layer between the views. In other words, it could develop units whose activation is dominated by a single view. This makes the MAE not suitable for transfer learning, since the views are not guaranteed to be projected to a common subspace. This is indeed verified by the results reported in~\cite{ngiam11} where they show that CCA performs better than deep MAE for the task of transfer learning.

% ------- CORR-NET, combination ---------------
As CCA and MAE based approaches %have their own advantages and disadvantages, 
%These two approaches 
appear to be complementary, % characteristics,
%On one hand, we have CCA and its variants which aim to produce correlated common representations but lack reconstruction capabilities. On the other hand, we have MAE which aims to do self-reconstruction and cross-reconstruction but does not guarantee correlated common representations.
several methods tried to combine them in one framework~\cite{wang15}. For example, a MAE based approach called Correlational Neural Network (CorrNet)~\cite{corrnet16} tried to explicitly maximize the correlation between the views when projecting them into the common subspace. 

\section{DEEP ARCHITECTURE}
\label{sec:arch}
We aim to solve two fundamental tasks for RGB-D images: semantic segmentation and depth prediction. We assume that we are given a training set of $N$ RGB-D images ($\bfx_i,\bfd_i$), $i=1,\ldots,N$. All images are assumed to be resized to width $W$ and height $H$.  Depth images are in HHA representation and have the same value range as RGB images, $\bfx_i,\bfd_i \in R^{H \times W \times 3}$. 
%Y_i \in L^{H \times W}$, for all $i=1,\ldots,M$
%having the same size $H \times W$, along with the ground-truth labeling ($Y_i$). 
%The pixels are assumed to be drawn as i.i.d. samples following a categorical distribution. Based on this assumption, we may define a CNN model to perform multinomial logistic regression.
%In the semantic segmentation setting, 
RGB images are annotated with $\bfy_i\in L^{H \times W}$, where $L$ is the label set, $L=\{1,\ldots,K\}$. 
%When the SS task is coupled with 
In the case of depth estimation, %the depth images are noisy or incomplete, and 
we assume to have additionally the ground true values $\bfd^*_i \in R^{H\times W}$.

We propose an architecture composed of two separate branches, one for each modality, which are consecutively fed into a common representation network. 
Two individual modality networks are of the encoder-decoder type, where the encoder applies dilated convolution to extract an informative feature map, while the decoder applies "atrous" convolution at multiple scales to encode contextual information and refine the segmentation boundaries.
%Two individual modality networks are of the encoder-decoder type, where the encoder applies "atrous" convolution at multiple scales to encode contextual information, while the decoder refines the segmentation results along the object boundaries. 
This choice is motivated by the recent success of the encoder-decoder architecture of DeepLab network~\cite{deeplab2}. It has been also used in FuseNet~\cite{hazirbas16} and SegNet~\cite{badrinarayanan2017segnet} and has showed good segmentation performance. 

%We apply an encoder-decoder network for each of the two modality networks (see Figure~\ref{fig:arch}). %Each encoder is a pre-trained classification network like ResNet101; 
Both RGB and depth encoders are initialized by the Resnet101 model trained on the COCO dataset. The encoders generate the feature maps, which the decoders use in "atrous" spatial pyramid pooling (ASPP) to robustly segment objects at multiple scales. 

%These two branches extract features from RGB and depth images. We note that the depth images are in HHA representation to have the same value range as color images.
%, i.e. into the interval of [0,255]. 

Feature maps generated by two modality branches are fed to the common representation network implemented in the form of a multi-view autoecoder~\cite{wang15}.
Unlike the conventional fusion of RGB and depth feature maps, the multi-view autoencoder allows to extract the shared representation from either one or two views.

%================================
\subsection{Training RGB network}
\label{ssec:train-rgb}

%We use the cross-entropy loss as the objective function for training the stream network. The loss is summed up over all the pixels in  a  mini-batch. To address the different class weight in the dataset, we use median frequency balancing~\cite{} where the weight assigned to a class in the loss function is the ratio of the median  of  class  frequencies  computed  on  the  entire  training  set divided by the class frequency. This implies that larger classes in the training set have a smaller weight and, viceversa, the smaller  classes have higher weights.

Our architecture enables two different settings. In the {\it semantic segmentation} (SS) setting, both RGB and depth views are jointly used to segment an image. In the case of {\it segmentation and depth estimation} (SS-D), we face the multi-task setting and expect the two views to achieve good results in both tasks.

We first proceed by training two individual modality branches
%stream networks 
(see the purple and blue streams in Figure~\ref{fig:arch}). Let $\theta^I$ and $\theta^D$ be parameters of RGB and depth networks, respectively. Let $\bfx^p_i=g^I(\bfx_i;\theta^I)$ be the feature map extracted from the last %(fully connected) 
layer of the RGB decoder when applied to image $\bfx_i$, $\bfx^p_i \in  R^{d \times H^{'} \times W^{'}}$. Analogously, let $\bfd^p_i=g^D(\bfd_i;\theta^D)$ be the feature map extracted from depth decoder when applied to the depth image $\bfd_i$, $\bfd^p_i \in R^{d \times H^{'} \times W^{'}}$. 

The network is trained in two stages. We first train the modality branches,  %networks
then we train the entire network. In the first stage, we place a randomly initialized softmax classification layer on top of $g^I$ and train the RGB network to minimize the semantic loss of the training data. The semantic loss $L_{rgb}^{ss}$ is defined as the cross-entropy loss
%/ multinomial logistic loss, 
%-----------------
\begin{equation}
L_{rgb}^{ss}= -\sum^N_{i=1} P(\bfy_i|\widehat \bfx_i),\label{eq:rgb-ss}
\end{equation}
%----------------
%where $\bfW^I$ are the weights of the softmax layer mapping from $g^I(\cdot)$ to $L^{H\times W}$,
where $\widehat \bfx_i$ is a pixel-wise prediction for image $\bfx_i$,
$\bfy_i$ is the ground truth labels, $P(\bfy|\bfx)=\sum_j{\rm log}\ p(y_j|x_j)$, and $p(y_j|x_j)$ is the probability of semantic label $y_j$ at pixel $j$.

The network is trained using the stochastic gradient descent on mini-batches of RGB images. After the convergence, all parameters $\theta^I$ of the RGB network are kept for the second stage, except 
%$\bfW^I$ 
the last layer which will be replaced by the common representation and reconstruction layer.

%=========================================
\subsection{Training depth network}
\label{ssec:train-depth}

Training the depth branch depends on the task. 
In the SS setting, the depth network is trained, similarly to the RGB network, to minimize semantic loss % on the training data:
%-------------------
%\begin{equation}
$L_{d}^{ss}= -\sum^N_{i=1} P(\bfy_i|\widehat \bfd_i),$
%\label{eq:d-ss}
%\end{equation}
%------------------
where $\hat \bfd_i$ is a prediction for depth image $\bfd_i$.

In the SS-D setting, we train the depth branch to minimize the regression loss on the training depth data.
\begin{comment}
%------------------
\begin{equation}
L_{d}^{ss}= -\sum^N_{i=1} S(\bfd^*_i,\widehat \bfd_i),\label{eq:d-d}
\end{equation}
%------------------
where $S(\bfd_i^*,\widehat \bfd_i)$ measures the difference between the ground truth $\bfd^*_i$ and depth prediction $\widehat \bfd_i$. 
\end{comment}
We tested several state-of-art proposals for the loss function $L_{d}^{d}$. One is the scale-invarient loss~\cite{eigen14}; it measures the relationships between points in the image irrespectively of the absolute values. 
%For the predicted depth map $\bfd$ and the ground truth $\bfd^*$, $S(\bfd,\bfd^*)=\sum_{i,j} ((log(\bfd_i)-log(\bfd_j))-(log(\bfd^*_i)-log(\bfd^*_j)))^2$, where pixels in both $\bfd$ and $\bfd^*$ are indexed by $i$. 
%----------------------
%smooth L1 loss or Huber loss%
We also considered the standard $L_2$ and {\it smoothed} $L_1$ loss~\cite{girshick15}. Less sensitive to outliers that the $L_2$ loss, smoothed $L_1$ loss is defined as 
%\begin{equation}
%S(\bfd_i^*,\bfd_i^r) = 
$L^{smo}_1(\bfd_i^*,\widehat \bfd_i)= \sum_{j} D (\bfd_{ij}^*-\widehat \bfd_{ij}),$
%\label{eq:smoothL1}
%\end{equation}
where
\begin{equation}
    D(x)=\left\{
        \begin{array}{ll}
          0.5 x^2,     \quad \quad \textnormal{if} \, |x| < 1 \\
          |x| - 0.5, \quad {\rm otherwise.}
        \end{array}
              \right.
\end{equation}
%\begin{equation}
%    \alpha(x)=\left\{
%        \begin{array}{ll}
%          0.5 (\bfd^*_i - \bfd^p_i)^2,     \quad \textnormal{if} \, |\bfd^*_i - \bfd^p_i| < 1 \\
%          |\bfd^*_i - \bfd^p_i| - 0.5, \quad {\rm otherwise.}
%        \end{array}
%              \right.
%\end{equation}

%________________________________________________________________________________________________________

%In our architecture, the key ingredient is the {\it common representation}, which combines the feature maps of the  RGB and depth branches.  The layer is implemented as element-wise summation. In FuseNet, we always insert the fusion layer after the CBR block. By making use of fusion the discontinuities of the features maps computed on the depth image are added into the RGB branch in order to enhance the RGB feature maps. As it can be observed in many cases, the features in the color domain and in the geometric domain complement each other. Based on this observation, we propose two fusion strategies: a) dense fusion (DF), where the fusion layer is added after each CBR block of the RGB branch. b) sparse fusion (SF), where the fusion layer is only inserted before each pooling. 
%These two strategies are illustrated in Figure 3.

%===========================================
\subsection{Common representation}
\label{ssec:deep}
Common representation network is implemented as a multi-view autoencoder~\cite{corrnet16,ngiam11}. It includes a hidden layer and an output layer. The input to the hidden layer is two feature maps $\bfx^{p}, \bfd^{p}$ from two modality branches. Similar to conventional autoencoders, the input and output layer has the same shape as the input, $d \times H'\times W'$, whereas the hidden layer is shaped as $k \times H'\times W'$, with $k$ being often smaller than $d$ (in Figure~\ref{fig:arch}, $d$=256 and $k$=128).

Given a two-view input $\bfz=(\bfx^{p},\bfd^{p})$, the hidden layer computes an encoded representation as the convolution
%-------------
%$$h(\bfz) = h(conv(\bfW_x, \bfx^{p}) + conv(\bfW_d, \bfd^{p}) + \bfb),$$
$$h(\bfz) = h(\bfW_x \circledast \bfx^{p} + \bfW_d \circledast \bfd^{p} + \bfb),$$
%-------------
where $\bfW_x$, $\bfW_d$ are projection weights, $\bfb$ is a bias vector, and $h$ is an activation function, such as {\it sigmoid} or {\it tanh}. 

The output layer tries to reconstruct $\bfz$ from this hidden representation $h(\bfz)$ by computing 
%------------
$$\bfz^r = g([\bfV_x \circledast h(\bfz), \bfV_d \circledast h(\bfz)] + \bfb_r),$$
%$$\bfz^r = g([conv(\bfV_x, h(\bfz)),conv(\bfV_d, h(\bfz))] + \bfb_r),$$
%------------
where $\bfV_x$, $\bfV_d$ are reconstruction weights, $\bfb_r$ is a output bias vector and $g$ is an activation function.

Given data $\{(\bfx^{p}_i,\bfd^{p}_i)\}_{i=1}^N$ from RGB and depth branches, the common representation is designed to minimize the self- and cross-reconstruction errors. The first is the 
%minimizes the error 
of reconstructing $\bfx^{r}_i$ from $\bfx^{p}_i$ and $\bfd^{r}_i$ from $\bfd^{p}_i$. The second one is the error of reconstructing $\bfx^{r}_i$ from $\bfd^{p}_i$ and $\bfd^{r}_i$ from $\bfx^{p}_i$.

To achieve this goal, we try to find the parameter values ${\theta^A} =\{\bfW_x, \bfW_d, \bfV_x, \bfV_d, \bfb, \bfb_r\}$ by minimizing the reconstruction loss function $L_{rec}$, defined as follows
%\begin{equation}
$$
\sum^N_{i=1} \{ L_r[\bfz_i, g(h(\bfz_i))] + L_r [\bfz_i,g(h(\bfx^p_i))] + L_r[\bfz_i, g(h(\bfd^p_i))] \},
%\label{eq:mae}
$$
%\end{equation}
\noindent 
where $L_r$ is the reconstruction error, $L_r(\bfx,\bfx')=||\bfx-\bfx'||^2$.

%===========================================
\subsection{End-to-end learning}
\label{ssec:full}
Common representation allows to obtain the reconstructed feature maps for both RGB and depth images as %$\bfx^r = 
$f(g(\bfx^p))$ and $f(g(\bfd^p))$. The entire set of network parameters is $\theta =\{\theta^I, \theta^D, \theta^A\}$. The objective function to minimize is then defined as  
\begin{equation}
{\cal L}= L^{ss}_{rgb} + L_{d} + \lambda L_{rec},
\label{eq:losses}
\end{equation}
\noindent
where the depth branch loss $L_{d}$ is $L^{ss}_{d}$ in the SS setting and $L^d_{d}$ in the SS-D setting; $\lambda$ is a scaling parameter for the reconstruction loss.
% remove eq references ~(\ref{eq:ss-d}) adn ~(\ref{eq:ddd-d})
In the above formulation, the semantic, depth and reconstruction losses are optimized jointly. 
%In the final stage of the optimization, the common representation will reconstruct the feature maps using one or both modalities.
%The optimization using SGD on minim-batches of image pairs $(\bfx_i, \bfd_i)$.
 
%The correlation term can included in the total objective function $L$, by interacting with the other terms, it makes sure that the hidden representations of the two views are highly correlated.

In addition to the common representation and view reconstruction, we also considered a possibility of maximizing the view correlation, as suggested in %DeepCCA~\cite{wang15} and
CorrNet~\cite{corrnet16}. In such a case, we try to maximize the correlation between the hidden representations of the two views. The correlation term can be included in the objective function $\cal L$, %by interacting with the other terms, 
it makes sure that the hidden representations of the two views are highly correlated.

\subsection{Training and optimization}
\label{ssec:train}
The architecture is implemented on the PyTorch framework. At the first stage, we train individual branches independently. In the SS setting, we train RGB and depth branches with segmentation labels, they are denoted RGB-SS and D-SS. 
%Both encoders are initiated with the dilated ResNet101 model pretrained on the COCO dataset. 
Each branch is trained for 20,000 iterations using SGD with momentum 0.9, batch size 24, and minimizing modality losses, $L^{ss}_{rgb}$ and $L_{d}^{ss}$. We retain the model parameters $\theta^I$ and $\theta^D$ for the second stage.

In the SS-D setting, we train the RGB branch with segmentation labels (RGB-SS) and the depth branch with depth ground truth (D-D). The D-D branch is trained using the scale irrelevant loss $L_2$ loss
or the smooth $L_1$ loss. % that resulted to be the most stable loss for our problem.)
We apply weight decay=0.0005 and the polynomial decay for the learning rate, with the base LR=0.0001 and power=0.9.

In the second stage, we train the entire network. 
We start with the two branch parameters $\theta^I$ and $\theta^D$ trained in the first stage, and refine them as well as the common representation network $\theta^A$ by minimizing the objective function $\cal L$ which combines semantic, depth and reconstruction losses.
%In the final stage of the optimization, the common representation will reconstruct the feature maps for both views. 
%and we add a final convolutional module, aka a classifier for the segmentation task, and a Regressor for the depth estimation task).
We fine-tune the network end-to-end with the Adam optimizer, but we freeze parameters of two modality encoders, it allows to speed-up the training without performance loss.
%We  and we need to consider also an additional loss that is responsible for the  and reconstruction task.

For the segmentation task, we add data augmentation, by flipping and randomly rotating input images on an angle between [-10, 10] degrees. RGB-D images to be augmented are selected randomly, but the augmentation is identical for both views.
%The RGB images get normalized to range [-0.5, +0.5] and the depth maps get normalized to range [0, 1]. %All inputs are resized to (512, 512).

%==============================================
%--------- EVALUATION --------------------
%=================================================
\section{EVALUATION}
\label{sec:evaluation}

We evaluated the proposed network on two publicly available RGB-D datasets: NYU depth dataset, 2nd version~\cite{silberman12} and SUN~\cite{song15}.
% category --> classes
%\begin{enumerate}
%\item 
NYU2 is a popular dataset, with 27 indoor categories but not all well represented. Following the publicly available split~\cite{silberman12}, 27 categories are reorganized, including 13 most common categories and an 'other' category for all remaining images. The training/test split is 795/654 images. Images are resized to $512 \times 512$ at training time, full size images are used at test time.
%\item 
SUN dataset contains 10,335 RGB-D images with 40 categories~\cite{song15}. Following the publicly available split with 37 most common and {\it other} categories~\cite{handa16}, it consists of 5,285 images for training and 5,050 images for testing. Images are resized to $360 \times 360$ at training time; full size images are used at test time. %The split is provided in the toolbox of SUN RGB-D dataset~\cite{song15}. 
All depth images are encoded using the HHA representation.
%We additionally consider a version with 37 categories.
%\end{enumerate}

%--------------------------
\begin{figure*}[th]
%{a) NUY2 RGB-D example.}
\centering{
\includegraphics[width=0.46\textwidth]{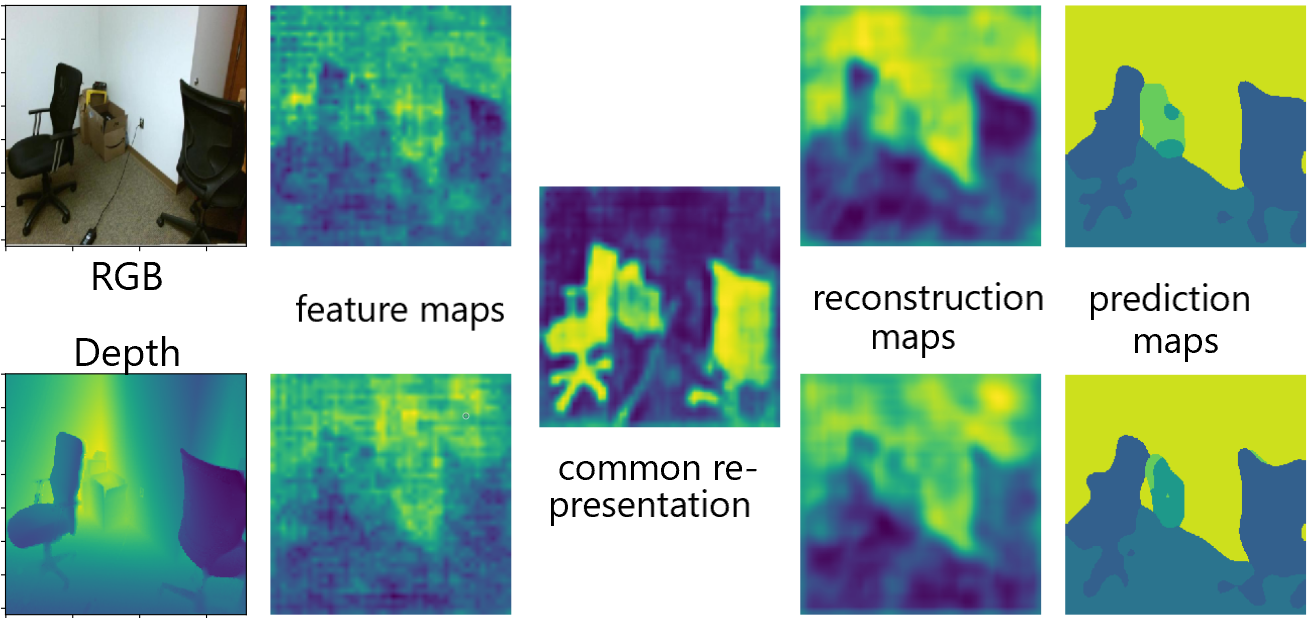}\quad \quad
\includegraphics[width=0.46\textwidth]{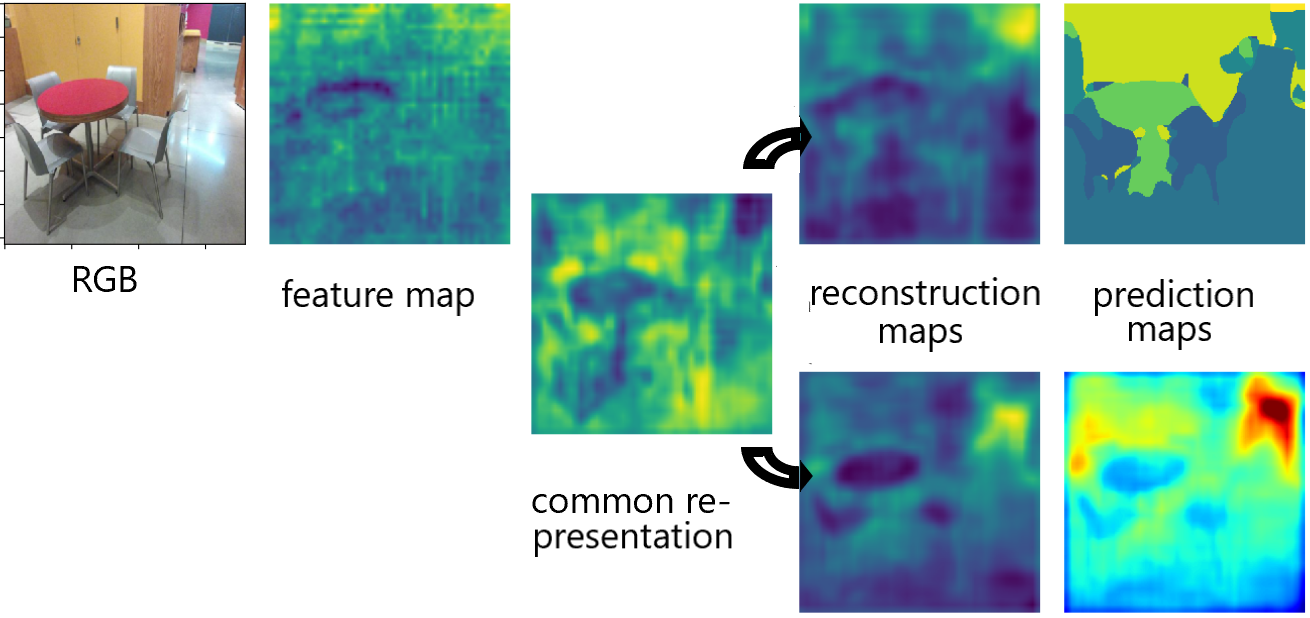}} %5-crop1.png and 2-crop1.png
%\centering{a) \quad\quad\quad \quad \quad \quad \quad \quad \quad b)}
%\centering{\includegraphics[width=1.0\textwidth]{figures/51-1-crop.png}}
%\centering{\includegraphics[width=1.0\textwidth]{figures/107-1-crop.png}}
%{b) SUN RGB-D example.}
\caption{left) Processing RGB-D images at different layers of the architecture. right) Reconstruction from one view.} 
\label{fig:featuremap}
\end{figure*}
%--------------------------

%==============================================
\subsection{Qualitative analysis}
\label{ssec:qualit}
We start with the qualitative analysis and test the proposed architecture with exemplar RGB-D images. Figure~\ref{fig:featuremap}.a shows how a NYU2 example gets processed by the network. In addition to the input images and ground truth segmentation, it shows feature maps extracted at different layers of the network. 
The upper row refers to the RGB branch, the lower row refers to depth branch. Column 2 visualizes feature maps generated by two modality decoders. Column 3 shows the common representations obtained from each modality map. A close resemblance of the two maps supports the concept of common representation which can be obtained from either view. Then, the reconstructed feature maps for both views are shown in column 4 and final predictions in column 5.

%--------------------------
%\begin{figure}[ht]
%\centering{\includegraphics[width=0.5\textwidth]{figures/2-crop1.png}} %cross-reconstruction1.png
%\caption{Example of the reconstruction from one view.} 
%\label{fig:cross}
%\end{figure}
%-------------------------

Figure~\ref{fig:featuremap}.b shows the cross-view reconstruction, where the RGB image is only available at test time. It starts with feature maps extracted from RGB network and the common representation. Then it shows how the common representation is used for two reconstruction and prediction maps.

%==============================================
\subsection{Quantitative results}
\label{ssec:eval}
%----------------------------
{\it Modularity.}
% to validate the common representation concept when applied to RGB-D data. 
% improve the encode-decoder branches.
% for semantic segmentation and depth estimation
The proposed architecture is designed in a {\it modular} way.
% with the principle of modularity in mind. \
It does not use any particular techniques invented to improve the semantic segmentation in special cases and image regions, such as multi-scale input, CRF, overlapping windows and cross-view ambiguities~\cite{lin2017refinenet,jiang17}. This choice is motivated by two main reasons. First, we wanted to test the effectiveness of the CRL in isolation, by excluding any impact of the additional improvements and the comparing to the state of art baseline. We paid most attention to the multi-view autoencoder and its capacity to generate common representation and reconstruct RGB and depth views.   
% by side effects not related with the main end-to-end training;
%
Second, the architecture design is general enough to cope with two different settings (SS and SS-D) and different modalities. We preferred the ability to work multi-modal and multi-task to an architecture narrowed to processing one particular task.
% clearly show how our method can be easily customized and adapted to, and not be constrained to particular learning configuration forced by network architectures or pre/post-processing routines. 
%We want to show how the learning representation and the network capacity are more important 
%than complex architecture design 
%in a general multi-modal and multi-task scenario. 
%And we want also to exploit transfer learning to obtain good results with a small amount of data and computational resources. 
Moreover, since the common representation is complementary to many of the state of art improvements, the proposed architecture can integrate most of them to boost the performance. 

%------------------------------------
{\it Evaluation metrics.}
%Two common criteria 
To evaluate our network on the segmentation task, we prefer 
%are the pixel accuracy and 
the intersection-over-union (IoU) score to the pixel accuracy. 
The pixel accuracy is known for being sensitive to the class disbalance, when many images include the large objects such as bed, wall, floor, etc.. The accuracy value may be misleading when the network performs better on the large objects and worse on the small ones. Instead, IoU score remains informative on both balanced and unbalanced datasets.

Let $C_{ij}$ denote the number of pixels those are predicted as class $j$ but actually belongs to class $i$, where $i,j \in L$. Then $C_{ii}$ denotes the number of pixels with correct prediction of class $i$. Let $T_i$ denote the total number of pixels that belongs to class $i$ in the ground truth, $K$ is the total number of classes in the dataset. Then
%---
%\begin{enumerate}
%\item {\it Pixel accuracy} measures the percentage of correctly classified pixels: 
%$Acc =\frac{\sum_i C_{ii}}{\sum_i T_i}$.
%\item {\it Intersection-over-union} 
IoU is the average value of the intersection between the ground truth and the predictions:
$IoU =\frac{1}{K} \sum_i \frac{C_{ii}}{T_i + \sum_j C_{ji} - C_{ii}}.$
%\end{enumerate}
%---

For depth estimation, we use the root mean square error (RMSE) that measures the error between the estimated depth and ground truth. 

%NYUv2 : (3, 512, 512) ---> (256, 65, 65)
%SUNRGBD : (3, 360, 360) ---> (256, 46, 46)
%===================================================
{\it Hyper-parameters.}  
We set $W^{'}=H^{'}=65$ for the NYU2 set and $W^{'}=H^{'}=46$ for the SUN set. Feature maps generated by modality branches are shaped with $d=256$. The number of hidden variables in the common representation is fixed, $k=128$. During the training with the objective function $\cal L$, weight $\lambda$ of reconstruction loss is 1.
%{\it We also trained the common representation network with different fractions of one and two view data.(check).}
%The reconstruction loss $L_{rec}$ in (\ref{eq:loss-rec}) is composed of three terms which refer to 
% the from original x from the and the 
%\end{itemize}

%===================================================
\subsection{Semantic segmentation and depth estimation}
\label{ssec:exp-ss}
 
%In the multi-view approach, 
We consider three different ways to use the architecture presented in Section~\ref{sec:arch} to process the RGB-D data. 

\begin{itemize}
\item \textbf{Independent learning}: In this case, each modality branch is trained and tested independently. In the SS setting, we have RGB-SS and D-SS branches; in the SS-D setting, we train and test RGB-SS and D-D branches to provide the baseline performances.
%A simple concatenation is included as the baseline.
\item \textbf{Joint learning}: The network is trained in two stages as described in Section~\ref{sec:arch}. In SS and SS-D settings, the network is trained to minimize the objective function $\cal L$,
with the appropriate depth branch loss $L_d$ (see Section 3.4).
%we test different scenarios for the SS and SS-D tasks
In either case, we compare them to the baselines obtained with the independent training. 
%\begin{itemize}
%\item 
In the {\bf SS} %is a {\it multi-view one-task} 
setting, we test the common representation with
one or two modalities available at test time,
%the self-, cross- and both view reconstructions, 
where the semantic segmentation is evaluated using the RGB, depth or both images.
%\item 
In the {\bf SS-D} %is a {\it multi-view multi-task} 
setting, we test the semantic segmentation and depth estimation with one or two modalities available at test time.
%with same, cross and both view reconstructions for the two tasks. % where the semantic segmentation 
%are evaluated using the RGB, depth or both images.
%\end{itemize}
\end{itemize}

Table I %~\ref{tab:nyu2} 
%evaluation results and compare the proposed architecture to the  the baselines. The table shows 
reports IoU values for the SS and SS-D settings on NYU2 dataset.
In the SS setting, training two modality branches independently yields 53.1 (RGB) and 37.1 (depth) IoU values; this reflects the RGB view being more informative than depth. Learning a joint model and using the common representation at test time improves the performance in cases when depth or both views are available.
As both views address the segmentation task, the common representation makes performance dependent on which view is available at test time. Instead it does not depend which view is being reconstructed.

In the SS-D setting, the baseline for RGB-SS branch is the same, the baseline for depth reconstruction using D-D branch gives RMSE value of 0.51. The common representation improves the RGB value to 54.3, and reduces the reconstruction error to 0.39 and 0.53 when using the depth only or both views, respectively. In the cross-view reconstruction, using depth for segmentation drops IoU value to 35.0 only, using RGB for depth estimation yields 0.72 error.

\begin{table}[th]
\centering
\begin{tabular}{||l||c||c|c||r|r|r||}  \hline
Setting &Branch & \multicolumn{2}{c||}{Independent} &\multicolumn{3}{c||}{Common representation}\\ 
\cline{3-7} 
        &       &  RGB      & Depth    & RGB   & Depth     & RGB+D   \\ \hline
      \multicolumn{7}{c}{NYU2 dataset} \\ \hline
SS      &RGB-SS & 53.1  & -     & 54.1 & 41.2      & \bf{57.6} \\
        &D-SS   &  -    & 37.1  & 54.2 & \bf{41.1} & 57.7      \\ \hline
SS-D    &RGB-SS & 53.1  & -     & 54.3 & 35.0      & \bf{55.2} \\
        &D-D    &    -   & 0.51 & 0.72 & \bf{0.39} & 0.53      \\ \hline 
      \multicolumn{7}{c}{SUN dataset} \\ \hline
SS      &RGB-SS & 39.7 & -     & 39.4  & 31.1      & \bf{42.4} \\
        &D-SS   &	 - & 31.1  & 39.4  & 31.1      & 42.3      \\ \hline
SS-D    &RGB-SS & 39.7 & -     & 39.3  & 20.3      & \bf{39.9} \\
        &D-D    & -    & 0.36  & 0.62  & \bf{0.31} & \bf{0.31} \\ \hline
\end{tabular}
\label{tab:sun37}
\caption{Independent and joint learning with one or two views at test time. The best results are shown in bold. %, improvements are underlined.
}
\end{table}

%--------------------- STATE OF ART COMPARISON -----------------

%\subsection{Comparison to the state of art}
%\label{ssec:soa}
 
\begin{comment}\begin{table}[th]
\centering
\begin{tabular}{||l||r|r|r||r|r|r||}  \hline
Methods     &\multicolumn{3}{c||}{Semantic Segmentation} &\multicolumn{3}{c||}{Depth Estimation} \\ \cline{2-7} 
%                          & RGB only & D only & RGB+Depth& RGB only & Depth only & RGB and Depth \\ \hline                                
                          & RGB      & D      & RGB+D    & RGB      & D          & RGB+D        \\ \hline                         
Baselines                 & 57.37    & 42.52  &		     &	        & 0.39       &              \\
Our method                & 57.59    & 47.73  & 60.81    &   0.61   & 0.32       &    0.32      \\ \hline
\cite{}                   &          &        &	         &	        &            &              \\
\cite{}                   &          &        &	         &	        &            &              \\ \hline
\end{tabular}
\label{tab:sun13}
\caption{SUN-13 set: Comparison on six different tasks.}
\end{table}
\end{comment}

%---------------------------------
\begin{table}[th]
\centering
\begin{tabular}{||l||r|r|r||r||}  \hline
Methods                    &\multicolumn{3}{c||}{Sem. Segmentation} &Depth Est. \\ \cline{2-5} 
                                              & RGB     & D     & RGB+D & RGB   \\ \hline      
                    \multicolumn{5}{c}{NYU2 dataset} \\ \hline
Our method                                    &\bf{54.1}&\bf{41.2}& 57.6  & 0.72  \\ \hline
Li et al.~\cite{li2015depth}                  &  -      &  -    &  -    & 0.82  \\
Roy et al.~\cite{roy2016monocular}            &  -      &  -    &  -    & 0.74  \\
Laina et al.~\cite{laina2016deeper}           &  -      &  -    &  -    &{\bf 0.57}  \\
Eigen et al.~\cite{eigen2015predicting}       &  -      &  -    & 52.6  & 0.64  \\
FuseNet-SF3~\cite{hazirbas16}                 &  -      &  -    & 56.0  &    -  \\
MVCNet-MaxPool~\cite{ma17}                    &  -      &  -    &\bf{ 59.0} &	 -  \\ \hline
%\end{tabular}
%\label{tab:nyu13}
%\caption{NYU2 set: Comparison on different tasks.}
%\end{table}
%---------------------------------
%\begin{table}[th]
%\centering
%\begin{tabular}{||l||r|r|r|r|r|r||}  \hline
%Methods                        &\multicolumn{3}{c|}{Semantic Segmentation} & Depth Est. \\ \cline{2-5} 
%& RGB      & D      & RGB+D    & RGB   \\ \hline   
                   \multicolumn{5}{c}{SUN dataset} \\ \hline                                           
Our method                                  & 39.49    &\bf{31.1} & 42.4     &\bf{0.62}  \\ \hline
Segnet \cite{badrinarayanan2017segnet}                              & 22.1     & -      &    -      & -     \\
Bayesian-Segnet \cite{kendall2015bayesian}  & 30.7     & -      &  -        & -     \\
Hazirbas et Ma \cite{hazirbas16}            & 32.4     & 28.8   & 33.6     & -     \\ 
FuseNet-SF5    \cite{hazirbas16}            & -        & -      & 37.3     & -     \\
DFCN-DCRF      \cite{jiang2017incorporating}& -        & -      & 39.3     & -     \\
Context-CRF    \cite{shen2017semantic}      & 42.3     & -      & -        & -     \\
RefineNet      \cite{lin2017refinenet}      &\bf{45.9}& -      & -        & -     \\
CFN            \cite{lin2017cascaded}       & -        & -      &\bf{ 48.1}   & -	   \\ \hline
%\cite{mausivian16}, 40cats&   &          & 39.3   &	              \\
%\cite{song17}                 &          &        & 42.4     &	      \\
%\cite{wang16}                 &          &        & 37.7     &	      \\
%\cite{jiang18}                              &          &        & 47.8     &	      \\ \hline
\end{tabular}
\label{tab:sun37}
\caption{Comparison to the state of art on different tasks.}
\end{table}
%--------------------------------

%===========================================
%\it{SUN dataset}
%\label{ssec:eval-sun}
Table I  %~\ref{tab:sun13} and~\ref{tab:sun37} 
also reports evaluation results on SUN dataset. Using both views does improve the performance, moreover depth estimation benefits more from the common representation than the segmentation task. %the cross-reconstruction resists well to the missing view.

We compare our results to the state of art on four typical scenarios for RGB-D images (see Table II). Our architecture is the only one able to cope with in all cases. moreover it remains competitive to the highly specialized architectures~\cite{li2015depth,jiang17} which cope with one or two scenarios only.

%====================================
\subsection{Discussion}
\label{ssec:disc}
Both quantitative and qualitative results validated the effectiveness of learning the common representation from RGB and depth images. However the conducted experiments left some questions open; we discuss them in this section.

In addition to the results reported in Tables I and II, %~\ref{tab:nyu2}-\ref{tab:sun37}, 
we tested a number of alternatives and made some conclusions. First, adding the view correlation (see Section~\ref{ssec:full}) does not seem to improve the common representation nor the performance. 
%in frozen and end-to-end training models. %Second, increasing the hidden space size (d) is likely to improve 
Second, the scale-irrelevant loss for the depth estimation, mentioned in Section~\ref{ssec:train-depth}, does not seem to perform better than the standard $L_2$ and smoothed $L_1$ losses; all SS-D results in table I and II refer to the smoothed $L_1$ loss.

The two-stage training of the network enables to play with a so-called frozen configuration. The modality branches trained at the first stage get frozen and extract feature maps for all RGB-D images in the dataset. Such a frozen configuration allowed to test different configurations of common representation network before training the full network end-to-end. 
%Modularity principle. 
%This allowed to select most promising configurations before the second, joint training phase. We refer to the frozen version of the our architecture after training and with .
%Below are some options which have been found promising in the frozen case, but not been tested in the end-to-end training yet.
Below we finally mention some ideas on further improving the current architecture.
%------------------------
\begin{enumerate}
%\item As mentioned in the introduction, the common representation allows to train the network with additional one view data. This composition has been successfully tested on the frozen configuration but still have to show the benefit in the full network training. 
\item The common representation is currently limited to one hidden layer. Using deeper multi-view autoencoders has been beneficial in the frozen case.
\item Learning the common representation is implemented on one fixed scale ($k=128$) of the RGB and depth feature maps. We consider replacing one-fixed-scale MAE with multi-scale ones, on each level of the encoder-decoder networks.
%has shown a clear advantage in the frozen scenario.
\item ResNet101 model pre-trained in COCO dataset fits well the segmentation task, but to the less extend the depth estimation task. We consider setting up a more appropriate pre-trained model or an option of training it from scratch or combine the two models~\cite{jiang17}.
%\item Finally we intend to test depth completion scenario  like toa dda the has not test  
\end{enumerate}
%--------------------------------------------

%=================================================
\section{CONCLUSION}
\label{sec:conclusion}
We proposed a new deep learning architecture for the  tasks of semantic segmentation and depth prediction from RGB-D images. In the proposed architecture, the conventional feature fusion is replaced with a common deep representation of the RGB and depth views. Combined with an encoder-decoder type of the network, the architecture allows for a joint learning for the semantic segmentation and depth estimation based on their common representation. This approach offers several important advantages, such as using one modality at test time to build a common representation and to reconstruct the missing modality. 
%In particular, this allows the cross-modality scenarios, such as getting an accurate semantically segmentation from the depth data only.
We reported a number of evaluation results on two standard RGB-D datasets.
 % showing the capacity to accurate segment the visual scene and predict the depth from one view only.
Both quantitative and qualitative results validated the effectiveness of learning the common representation from RGB and depth images. 
 
 \clearpage
%================================================
\bibliographystyle{plain} %{splncs}
\bibliography{semsegmentation}
%\printbibliography

\end{document}